\documentclass[10pt,twocolumn,letterpaper]{article}

\usepackage{cvpr}              

\usepackage{multirow}
\usepackage{array}
\usepackage{pifont}
\usepackage{tikz}
\usepackage{pgfplots}
\pgfplotsset{compat=1.18}
\usepgfplotslibrary{groupplots}
\usepackage{threeparttable}
\usepackage[accsupp]{axessibility}  
\usepackage[ruled]{algorithm2e}
\usetikzlibrary{shapes,arrows,arrows.meta,positioning,fit,backgrounds,calc,decorations.pathreplacing}

\newcolumntype{R}[1]{>{\raggedright\arraybackslash}p{#1}}

\definecolor{cvprblue}{rgb}{0.21,0.49,0.74}
\usepackage[breaklinks,colorlinks,allcolors=cvprblue]{hyperref}

\def\paperID{11}
\def\confName{CVPR}
\def\confYear{2026}

\title{A Survey of Spatial Memory Representations\\for Efficient Robot Navigation}

\author{Ma.\ Madecheen S.\ Pangaliman$^{1,2}$ \quad Steven S.\ Sison$^{1}$ \quad Erwin P.\ Quilloy$^{1}$ \quad Rowel O. Atienza$^{1}$\\
$^{1}$University of the Philippines Diliman \quad $^{2}$University of Santo Tomas}  

\begin{document}
\maketitle

\begin{abstract}
As vision-based robots navigate larger environments, their spatial memory grows without bound, eventually exhausting computational resources, particularly on embedded platforms (8--16\,GB shared memory, $<$30\,W) where adding hardware is not an option. This survey examines the spatial memory efficiency problem across 88 references spanning 52 systems (1989--2025), from occupancy grids to neural implicit representations. We introduce the \textit{overhead factor} $\alpha = M_{\text{peak}} / M_{\text{map}}$, the ratio of peak runtime memory (the total RAM or GPU memory consumed during operation) to saved map size (the persistent checkpoint written to disk), exposing the gap between published map sizes and actual deployment cost. Independent profiling on an NVIDIA A100 GPU reveals that $\alpha$ spans two orders of magnitude within neural methods alone, ranging from 2.3 (Point-SLAM) to 215 (NICE-SLAM, whose 47\,MB map requires 10\,GB at runtime), showing that memory architecture, not paradigm label, determines deployment feasibility. We propose a standardized evaluation protocol comprising memory growth rate, query latency, memory--completeness curves, and throughput degradation, none of which current benchmarks capture. Through a Pareto frontier analysis with explicit benchmark separation, we show that no single paradigm dominates within its evaluation regime: 3DGS methods achieve the best absolute accuracy at 90--254\,MB map size on Replica, while scene graphs provide semantic abstraction at predictable cost. We provide the first independently measured $\alpha$ reference values and an $\alpha$-aware budgeting algorithm enabling practitioners to assess deployment feasibility on target hardware prior to implementation.
\end{abstract}

\begin{figure}[t]
\centering
\resizebox{\columnwidth}{!}{%
\begin{tikzpicture}
    \draw[thick, -latex] (0,0) -- (9.5,0) node[right, font=\normalsize] {Memory Efficiency};
    \draw[thick, -latex] (0,0) -- (0,8.0) node[above, font=\normalsize, yshift=2pt] {\rotatebox{0}{Geometric Completeness}};
    \node[font=\small, below] at (1.0,0) {Low};
    \node[font=\small, below] at (8.5,0) {High};
    \node[font=\small, left] at (0,1.0) {Low};
    \node[font=\small, left] at (0,7.0) {High};

    \node[star, star points=5, star point ratio=2.25, fill=yellow!70, draw=orange, minimum size=10pt, inner sep=0pt] at (8.8,7.5) {};
    \node[font=\small\bfseries, orange!80!black, left] at (8.6,7.5) {Ideal};

    \node[draw=blue!60, fill=blue!12, thick, rounded corners=4pt,
          minimum width=2.8cm, minimum height=1.3cm,
          align=center, font=\small] (dense) at (2.0,6.5)
          {\textbf{Dense Grids}\\[-1pt]{\footnotesize GBs / building}};

    \node[draw=orange!70, fill=orange!12, thick, rounded corners=4pt,
          minimum width=2.8cm, minimum height=1.3cm,
          align=center, font=\small] (sparse) at (4.5,2.0)
          {\textbf{Sparse Features}\\[-1pt]{\footnotesize 55\,MB, no geometry}};

    \node[draw=purple!60, fill=purple!10, thick, rounded corners=4pt,
          minimum width=3.0cm, minimum height=1.3cm,
          align=center, font=\small] (neural) at (5.0,5.5)
          {\textbf{Neural Implicit}\\[-1pt]{\footnotesize 8\,MB map; 1.3\,GB GPU}};

    \node[draw=green!60!black, fill=green!10, thick, rounded corners=4pt,
          minimum width=2.8cm, minimum height=1.3cm,
          align=center, font=\small, densely dashed] (scene) at (8.0,4.0)
          {\textbf{Scene Graphs}\\[-1pt]{\footnotesize 48\,MB graph layer}};

    \draw[green!60!black, dashed, thick] (scene.south) -- +(0,-1.0)
          node[below, font=\footnotesize, align=center, text=green!40!black, draw=none]
          {(graph layer only)\\[1pt]+metric backend (scene-dep.)};

    \draw[-latex, very thick, blue!40, shorten >=3pt, shorten <=3pt]
        (dense.south) -- (2.0,2.0) -- (sparse.west);
    \node[circle, fill=blue!40, text=white, font=\footnotesize\bfseries, inner sep=2pt] at (2.0,4.2) {1};

    \draw[-latex, very thick, purple!40, shorten >=3pt, shorten <=3pt]
        (sparse.north) -- (4.5,3.8) -- (neural.south);
    \node[circle, fill=purple!40, text=white, font=\footnotesize\bfseries, inner sep=2pt] at (4.5,3.4) {2};

    \draw[-latex, very thick, green!50!black, shorten >=3pt, shorten <=3pt]
        (neural.east) -- (8.0,5.5) -- (scene.north);
    \node[circle, fill=green!50!black, text=white, font=\footnotesize\bfseries, inner sep=2pt] at (6.8,5.5) {3};

    \node[circle, fill=blue!40, text=white, font=\footnotesize\bfseries, inner sep=2pt] at (0.5,1.2) {1};
    \node[anchor=west, draw=none, font=\footnotesize, inner sep=0pt] at (0.8,1.2) {1990s--2010s};
    \node[circle, fill=purple!40, text=white, font=\footnotesize\bfseries, inner sep=2pt] at (0.5,0.8) {2};
    \node[anchor=west, draw=none, font=\footnotesize, inner sep=0pt] at (0.8,0.8) {2020--2023};
    \node[circle, fill=green!50!black, text=white, font=\footnotesize\bfseries, inner sep=2pt] at (0.5,0.4) {3};
    \node[anchor=west, draw=none, font=\footnotesize, inner sep=0pt] at (0.8,0.4) {2023--2025};
\end{tikzpicture}%
}
\caption{Evolution of spatial memory representations along two competing demands: geometric completeness and memory efficiency. Sparse features~(\textbf{1}) traded completeness for efficiency; neural methods~(\textbf{2}) recovered completeness via learned compression; scene graphs~(\textbf{3}) added semantic abstraction. The scene graph box is dashed to indicate that the 48\,MB figure reflects \textit{only} the graph abstraction layer~\cite{hughes2022hydra}; the required metric-semantic backend~\cite{rosinol2020kimera} adds scene-dependent cost (dashed arrow), shifting the true system footprint leftward. The scene graph's vertical position reflects its semantic completeness (objects, rooms, places) rather than geometric fidelity.}
\label{fig:teaser}
\end{figure}
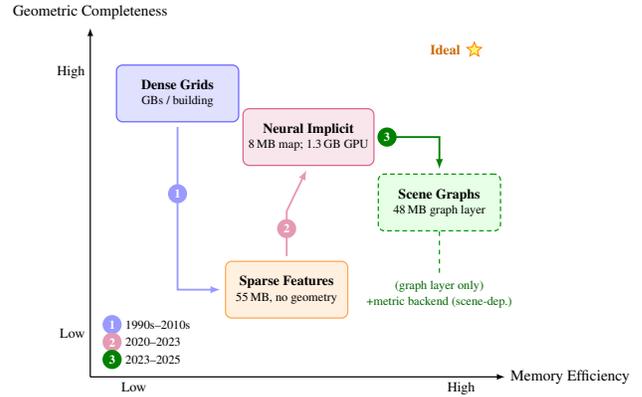

\begin{figure*}[t]
\centering
\resizebox{\textwidth}{!}{%
\begin{tikzpicture}[
    every node/.style={draw, rounded corners, align=center, font=\small, inner sep=4pt},
    edge/.style={draw, -latex},
    metricnode/.style={font=\scriptsize, inner sep=3pt},
    leaf/.style={font=\scriptsize, inner sep=3pt, fill=blue!5},
]
    \node (root) at (12,0) {Spatial Memory\\Representations};
    \node (metric) at (3,-2.4) {Metric Maps};
    \node (topo) at (10,-2.4) {Topological\\Maps};
    \node (sem) at (16,-2.4) {Semantic\\Maps};
    \node (hier) at (22,-2.4) {Hierarchical\\Representations};

    \node[metricnode] (occ) at (0,-5) {Occupancy Grids\\\cite{elfes1989using,hornung2013octomap}\\[-1pt]{\color{gray!40!black}\tiny $\sim$96\,MB/bldg}};
    \node[metricnode] (feat) at (3.5,-5) {Feature Maps\\\cite{mur2015orb,engel2014lsd}\\[-1pt]{\color{gray!40!black}\tiny ATE 3--11\,cm, 35--55\,MB}};
    \node[metricnode] (neur) at (7,-5) {Neural Implicit\\\cite{mildenhall2020nerf,kerbl20233d}\\[-1pt]{\color{gray!40!black}\tiny ATE 0.4--3.1\,cm, 1\,MB--2.9\,GB}};
    \node[metricnode] (place) at (9.5,-5) {Place Graphs\\\cite{cummins2008fabmap}\\[-1pt]{\color{gray!40!black}\tiny topology only}};
    \node[metricnode] (nav) at (12,-5) {Navigation\\Graphs \cite{shah2023lmnav}\\[-1pt]{\color{gray!40!black}\tiny $<$1\,MB graph}};
    \node[metricnode] (obj) at (14.5,-5) {Object Level\\\cite{rosinol2020kimera}\\[-1pt]{\color{gray!40!black}\tiny ATE $\sim$5\,cm, 48\,MB}};
    \node[metricnode] (dsem) at (17,-5) {Dense Semantic\\\cite{zhu2024snislam}\\[-1pt]{\color{gray!40!black}\tiny ATE 0.4\,cm, 92\,MB}};
    \node[metricnode] (openvoc) at (19.5,-5) {Open Vocab.\\\cite{gadre2023cows}\\[-1pt]{\color{gray!40!black}\tiny +2\,GB features}};
    \node[metricnode] (sg) at (21.5,-5) {Scene Graphs\\\cite{hughes2022hydra}\\[-1pt]{\color{gray!40!black}\tiny 48\,MB, multi-layer}};
    \node[metricnode] (ml) at (24,-5) {Multi-Level\\Maps \cite{tian2022kimeramulti}\\[-1pt]{\color{gray!40!black}\tiny hierarchical}};

    \node[leaf] (dense) at (-1,-8) {Dense\\\cite{elfes1989using}};
    \node[leaf] (octree) at (1,-8) {Octree\\\cite{hornung2013octomap}\\[-1pt]{\color{gray!40!black}\tiny $\sim$45\,MB/bldg}};
    \node[leaf] (sp) at (2.7,-8) {Sparse\\\cite{mur2015orb}};
    \node[leaf] (semid) at (4.5,-8) {Semi-dense\\\cite{engel2014lsd}};
    \node[leaf] (nerf) at (6.2,-8) {NeRF\\\cite{sucar2021imap}};
    \node[leaf] (gs) at (7.8,-8) {3DGS\\\cite{keetha2024splatam}};

    \draw[edge] (root) -- (metric);
    \draw[edge] (root) -- (topo);
    \draw[edge] (root) -- (sem);
    \draw[edge] (root) -- (hier);
    \draw[edge] (metric) -- (occ);
    \draw[edge] (metric) -- (feat);
    \draw[edge] (metric) -- (neur);
    \draw[edge] (topo) -- (place);
    \draw[edge] (topo) -- (nav);
    \draw[edge] (sem) -- (obj);
    \draw[edge] (sem) -- (dsem);
    \draw[edge] (sem) -- (openvoc);
    \draw[edge] (hier) -- (sg);
    \draw[edge] (hier) -- (ml);
    \draw[edge] (occ) -- (dense);
    \draw[edge] (occ) -- (octree);
    \draw[edge] (feat) -- (sp);
    \draw[edge] (feat) -- (semid);
    \draw[edge] (neur) -- (nerf);
    \draw[edge] (neur) -- (gs);
\end{tikzpicture}%
}
\caption{Taxonomy of spatial memory representations with representative citations and typical efficiency metrics (gray). ATE = absolute trajectory error on standard benchmarks; map sizes are representative for building-scale environments. 3DGS = 3D Gaussian Splatting. Semantic maps include open-vocabulary approaches leveraging vision-language models such as CLIP.}
\label{fig:taxonomy}
\end{figure*}
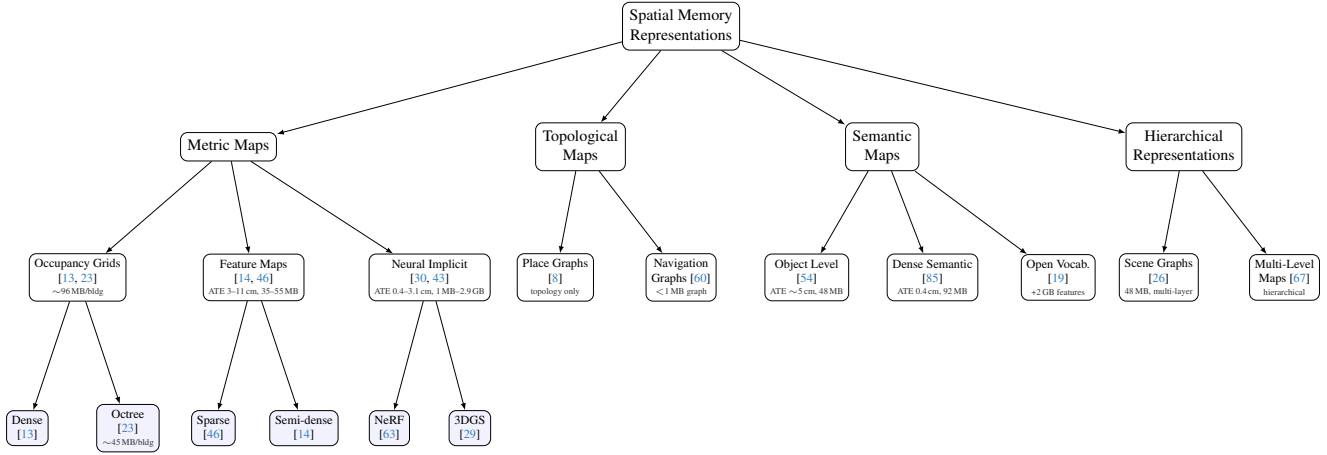

\section{Introduction}
\label{sec:intro}

Visual simultaneous localization and mapping (SLAM) has been one of the most actively studied problems in computer vision for over two decades~\cite{davison2003real,klein2007parallel}, underpinning applications from autonomous driving to augmented reality and domestic robotics. Despite substantial progress, a fundamental limitation emerges at deployment scale: ORB-SLAM3~\cite{campos2021orb} operates within a few hundred megabytes on a five-minute EuRoC~\cite{euroc} sequence, a standard visual-inertial benchmark comprising stereo-camera and IMU recordings in a machine hall and a Vicon room at ETH Zurich, yet mapping an entire office building (${\sim}3{,}000$\,m$^3$) over a workday strains available memory. For neural SLAM systems, a city-block-scale mission can exhaust GPU memory well before completion.

For readers entering this field, the core challenge is straightforward: a robot must simultaneously determine its position (\textit{localization}) and build a map of its surroundings (\textit{mapping}). The \textit{spatial memory} is the persistent data structure encoding this map. Different representations offer distinct tradeoffs between geometric detail, memory cost, and query speed (Fig.~\ref{fig:taxonomy}). This survey organizes these representations along the memory efficiency axis. Readers new to the field can use Fig.~\ref{fig:taxonomy} as a roadmap, Table~\ref{tab:selection_guide} to match a representation to hardware constraints, and Algorithm~\ref{alg:budgeting} to compute the maximum feasible map size.

This spatial memory growth problem, where representations expand continuously with environment size and mission duration, is the central focus of this survey (Fig.~\ref{fig:teaser}). The consequences compound: query latency exceeds real-time deadlines, map updates fall behind the sensor stream, and on resource-constrained platforms, exceeding available RAM terminates the system. Three root causes drive the growth: \textit{spatial extent} (dense representations scale as $O(V)$ with mapped volume), \textit{mission duration} (observations accumulate linearly with time), and \textit{revisitation patterns} (naive approaches store redundant data while more efficient designs exploit repetition for compression).

One might argue that memory is inexpensive and abundant, but deployment platforms present a different picture. Autonomous robots, drones, and AR headsets operate on embedded GPUs with 8--16\,GB shared memory (e.g., NVIDIA Jetson Orin) under strict power budgets ($<$30\,W), and upgrading hardware on a deployed robot is not feasible. SplaTAM ($\alpha_{\text{GPU}} = 55$) consuming 14\,GB at runtime leaves $<$2\,GB for perception, planning, and the OS, making the system infeasible despite the map being only 254\,MB. Scaling to a 100\,m$^2$ apartment would extrapolate to ${\sim}200$\,GB, far exceeding even a data-center-class A100 (80\,GB). Memory efficiency is therefore not a cost problem but a \textit{feasibility} constraint.

Recent surveys~\cite{tosi2024nerf3dgs,chen2024survey3dgs} organize neural SLAM systems by \textit{method} and benchmark accuracy. Our work is complementary: we organize systems by \textit{memory behavior}, treating neural methods as one instance of a broader memory-scaling problem. This reframing changes practical recommendations: Tosi et al.\ would rank Co-SLAM favorably based on its 8\,MB map, while our $\alpha$ analysis reveals that the same system requires 1.3\,GB at runtime ($\alpha_{\text{GPU}} = 157$).

\subsection{Contributions}

We make the following contributions:
\begin{enumerate}
\item We introduce the \textit{overhead factor} $\alpha = M_{\text{peak}} / M_{\text{map}}$, a diagnostic metric that distinguishes $\alpha_{\text{CPU}}$ (process Resident Set Size, RSS) from $\alpha_{\text{GPU}}$ (device allocation), and compile the first cross-paradigm comparison of runtime versus persistent memory, exposing the gap between published map sizes and actual deployment cost.
\item We independently profile five neural SLAM systems (Co-SLAM, NICE-SLAM, Point-SLAM, SplaTAM, SGS-SLAM) on an NVIDIA A100 GPU, revealing that $\alpha_{\text{GPU}}$ varies by two orders of magnitude even within neural methods: from 2.3 (Point-SLAM) to 215 (NICE-SLAM, whose 47\,MB map requires 10\,GB at runtime). These are, to our knowledge, the first independently measured $\alpha_{\text{GPU}}$ values for neural SLAM (Table~\ref{tab:efficiency}, Fig.~\ref{fig:memory_growth}).
\item We propose a standardized evaluation protocol with four complementary metrics beyond static map-size reporting (Section~\ref{sec:evaluation}), and an $\alpha$-aware budgeting algorithm that lets practitioners compute maximum feasible map size from two inputs, namely the memory budget and $\alpha$, before selecting a representation (Algorithm~\ref{alg:budgeting}).
\item We present a Pareto frontier analysis with explicit benchmark separation (Fig.~\ref{fig:pareto}), a taxonomy of 52 systems across four paradigm families (Fig.~\ref{fig:taxonomy}), memory dynamics categorization, and concrete open challenges with specific research questions.
\end{enumerate}

\section{Scope and Methodology}

This survey focuses on the \textit{representation layer} of \textit{vision-based} spatial memory: how spatial knowledge from cameras and IMUs is encoded, compressed, and managed over time. Core building blocks (ORB~\cite{rublee2011orb}, SuperPoint~\cite{detone2018superpoint}, NeRF~\cite{mildenhall2020nerf}, 3DGS~\cite{kerbl20233d}, CLIP) originate in mainstream computer vision, yet their downstream memory implications are rarely discussed. We exclude full autonomy stacks, perception without persistent memory, and LiDAR-based mapping (except BioSLAM~\cite{yin2023bioslam}). We prioritize breadth; for deeper NeRF/3DGS treatment see~\cite{tosi2024nerf3dgs,chen2024survey3dgs}.

We surveyed papers from 1989 to 2025, filtering for systems that introduce or evaluate persistent spatial memory representations. This yielded 88 references covering approximately 52 distinct systems (15 directly compared in efficiency tables); the remainder are datasets, benchmarks, optimization techniques, and related surveys. Section~\ref{sec:representations} covers all four representation families (including hierarchical and semantic maps); Section~\ref{sec:efficiency} presents our unified efficiency analysis; Section~\ref{sec:evaluation} combines the evaluation protocol with practical deployment guidance; Section~\ref{sec:dynamics} categorizes memory dynamics; and Section~\ref{sec:challenges} identifies open problems.

\section{Spatial Memory Representations: A Memory-Centric Overview}
\label{sec:representations}

We review the four dominant representation families through the lens of memory efficiency: scaling, $\alpha$, and compression strategies. Tables~\ref{tab:feature_slam} and~\ref{tab:nerf_slam} collect comparable metrics; for system-level SLAM contributions see~\cite{tosi2024nerf3dgs,chen2024survey3dgs}.

\subsection{Dense Maps: \texorpdfstring{$O(V)$}{O(V)} Scaling}
\label{sec:dense}

Occupancy grids~\cite{elfes1989using} scale linearly with mapped volume:
\begin{equation}
M_{\text{grid}} = \frac{V}{r^3} \times b
\label{eq:dense_scaling}
\end{equation}
where $V$ is mapped volume, $r$ is voxel resolution, and $b$ is bytes per cell. At 5\,cm resolution with $b{=}4$\,B, a 3{,}000\,m$^3$ building floor requires ${\sim}96$\,MB, which is already prohibitive for embedded platforms. OctoMap~\cite{hornung2013octomap} compresses this $2$--$13\times$ via octree subdivision (${\sim}45$\,MB at building scale; Table~\ref{tab:efficiency}); UFOMap~\cite{duberg2020ufomap} and Voxblox~\cite{oleynikova2017voxblox} (based on Truncated Signed Distance Fields, TSDF; see Table~\ref{tab:selection_guide}) offer further reductions, but the $O(V)$ scaling wall persists for any volumetric representation. No dense-only system reports $\alpha$; runtime overhead is dominated by sensor processing rather than map maintenance, so $\alpha$ is expected to be low ($<$10) but remains unverified.

\subsection{Sparse Features: Low Overhead, Limited Geometry}
\label{sec:sparse}

Feature-based SLAM~\cite{mur2015orb,mur2017orb,campos2021orb} trades dense geometry for an order-of-magnitude memory reduction, storing sparse keypoint landmarks, keyframes, and covisibility graphs. Semi-dense methods such as LSD-SLAM~\cite{engel2014lsd} reconstruct along image gradients, offering richer geometry at moderate memory increase but without the full $O(V)$ cost of dense grids. A typical building-scale ORB-SLAM3 session yields an estimated ${\sim}55$\,MB payload with $\alpha_{\text{CPU}} \approx 4$ (180--250\,MB peak CPU RSS on EuRoC~\cite{euroc}, Intel i7-10700). This low $\alpha$ means map size reliably predicts deployment cost. Visual-inertial systems exhibit similar profiles: ORB-SLAM3 achieves the best ATE (3.5\,cm) and the highest composite efficiency (Eff.~=~5.2) in Table~\ref{tab:feature_slam}, while Basalt~\cite{usenko2020basalt} trades accuracy (8.8\,cm, Eff.~=~3.2) for the smallest map (35\,MB) and lowest peak memory (120\,MB); VINS-Mono~\cite{qin2018vins} (10.6\,cm monocular-inertial, Eff.~=~2.4) trails on ATE but operates within 150\,MB peak using only a monocular camera. All three sparse systems sustain 20--30\,FPS (Table~\ref{tab:feature_slam}), comfortably real-time on a single CPU core.

Two memory management strategies are particularly instructive. RTAB-Map~\cite{labbe2019rtabmap} bounds active memory via a working memory / long-term memory transfer: infrequently accessed nodes are moved to disk regardless of session length. MS-SLAM~\cite{zhang2024msslam} applies sliding-window sparsification, reporting $>$70\% reduction in peak memory increase relative to ORB-SLAM3 while maintaining accuracy. Learned-flow systems (DROID-SLAM~\cite{teed2021droidslam}, 11\,GB GPU; DPVO~\cite{teed2024dpvo}) achieve strong accuracy but exhibit GPU-dominated memory profiles ($\alpha \gg 1$), presaging the overhead patterns observed in neural methods. Table~\ref{tab:feature_slam} compares systems with comparable memory metrics.

\begin{table}[t]
\caption{Feature-based visual SLAM comparison on EuRoC~\cite{euroc}
(MH01--MH05; sensing modes vary per system; see footnotes). Eff.\ = $10^3/(\text{ATE}\times M_{\text{map}})$; see Section~\ref{sec:efficiency} for caveats on this composite metric.}
\label{tab:feature_slam}
\centering
\scriptsize
\begin{threeparttable}
\begin{tabular*}{\columnwidth}{@{\extracolsep{\fill}}lccccc}
\toprule
\textbf{System} & \textbf{ATE} & \textbf{Map} & \textbf{Peak} & \textbf{FPS} & \textbf{Eff.} \\
 & (cm) & (MB) & (MB) & & \\
\midrule
VINS-Mono~\cite{qin2018vins} & 10.6\tnote{$\|$} & 40\tnote{$*$} & 150\tnote{$*$} & 20\tnote{$*$} & 2.4 \\
Basalt~\cite{usenko2020basalt} & 8.8\tnote{$\S$} & 35\tnote{$*$} & 120\tnote{$*$} & 30\tnote{$*$} & 3.2 \\
ORB-SLAM3~\cite{campos2021orb} & 3.5\tnote{$\dag$} & 55\tnote{$\ddag$} & 220\tnote{$\ddag$} & 30 & 5.2 \\
\bottomrule
\end{tabular*}
\begin{tablenotes}
\scriptsize
\item[$*$] Survey estimate from profiling; the cited paper
does not report this value directly.
\item[$\S$] Average RMS ATE over MH01--MH05 stereo
(proposed VIO); computed from Table~I
of~\cite{usenko2020basalt}.
\item[$\dag$] Average RMS ATE across all 11 EuRoC sequences
(stereo-inertial)~\cite{campos2021orb}. Note: the 9\,mm figure
sometimes associated with ORB-SLAM3 refers to TUM-VI, not EuRoC.
\item[$\|$] Average RMS ATE over MH01--MH05 (monocular-inertial);
reproduced by ORB-SLAM3 authors~\cite{campos2021orb}, Table~II.
\item[$\ddag$] Map and peak memory are survey estimates;
\cite{campos2021orb} does not report memory usage.
\end{tablenotes}
\end{threeparttable}
\end{table}

\subsection{Neural Maps: High Overhead, Hidden Cost}
\label{sec:neural}

Neural representations learn compact functions~\cite{mildenhall2020nerf}:
\[
F_\theta: (\mathbf{x}, \mathbf{d}) \rightarrow (\mathbf{c}, \sigma)
\]
Maps are remarkably small (iMAP~\cite{sucar2021imap}: ${\sim}1$\,MB; Co-SLAM~\cite{wang2023coslam}: 8\,MB), but runtime GPU memory reveals the true cost: Co-SLAM requires 1.3\,GB ($\alpha_{\text{GPU}} = 157$) due to optimizer state, gradient buffers, and rendering caches (analyzed in Section~\ref{sec:efficiency}). In contrast, iMAP's tiny map yields the highest within-paradigm composite efficiency (Eff.~=~320.5, Table~\ref{tab:nerf_slam}), despite having the worst ATE (3.12\,cm) and PSNR (22.1\,dB); at the other extreme, Point-SLAM's 2.9\,GB map yields Eff.~=~0.7, a $458\times$ spread illustrating why we advocate Pareto analysis over scalar ratios. Throughput also varies widely: Co-SLAM reaches 16\,FPS, while Point-SLAM runs at only 1\,FPS and 3DGS systems range from 2\,FPS (SGS-SLAM) to 8\,FPS (GS-SLAM; Table~\ref{tab:efficiency}). Hash-based methods~\cite{wang2023coslam,johari2023eslam,mueller2022instantngp} bound $M_{\text{map}}$ by construction, but quality degrades silently when scene complexity approaches table capacity, providing bounded memory rather than constant-quality scaling.

3D Gaussian Splatting (3DGS)~\cite{kerbl20233d} stores explicit primitives (${\sim}236$\,B each) scaling as $O(N_G)$ with scene surface area; without management, growth is unbounded. Compression is an active research area: learned masking with codebook quantization~\cite{lee2024compact} achieves $10$--$25\times$, pruning with vector quantization~\cite{fan2024lightgaussian} ${\sim}15\times$, and Gaussian merging~\cite{bai2025memgs} further reduces count. Parallel work extends 3DGS SLAM along robustness~\cite{zheng2025wildgsslam}, loop closure~\cite{zhu2024loopsplat}, and semantics~\cite{li2024sgsslam,katragadda2025onlinelang} axes, but critically none of the recent systems~\cite{huang2024photoslam,sandstrom2024splat,ha2025rgbdgsicpslam,peng2024rtgslam,chen2025splatnav,lei2025gaussnav,wen2025segsslam,gao2025gevo} report $M_{\text{peak}}$ or $\alpha$, leaving deployment costs unknown. The memory challenge differs by architecture: compact implicit methods and semantic 3DGS exhibit large gaps between map and runtime cost ($\alpha_{\text{GPU}} = 55$--$215$ for measured systems; SGS-SLAM's $\alpha = 159$ rivals hash-based Co-SLAM), while dense neural point methods shift the burden to the map itself (Point-SLAM: 2.9\,GB, $\alpha = 2.3$; Table~\ref{tab:efficiency}). Neither regime achieves both compact maps and low runtime overhead.

\begin{table}[t]
\caption{Neural SLAM on Replica (synthetic). Eff.\ = $10^3/(\text{ATE}\times M_{\text{map}})$, higher is better. $\dagger$: CVPR/ECCV 2024 or later.}
\label{tab:nerf_slam}
\centering
\scriptsize
\begin{threeparttable}
\begin{tabular*}{\columnwidth}{@{\extracolsep{\fill}}lccccc}
\toprule
\textbf{System} & \textbf{ATE} & \textbf{PSNR} & \textbf{Map} & \textbf{Peak} & \textbf{Eff.} \\
 & (cm) & (dB) & (MB) & (MB) & \\
\midrule
iMAP~\cite{sucar2021imap} & 3.12\tnote{f} & 22.1\tnote{d} & 1\tnote{a} & ---\tnote{d} & 320.5 \\
NICE-SLAM~\cite{zhu2022nice} & 1.06\tnote{b} & 24.4\tnote{g} & 47\tnote{e} & 10,082\tnote{e} & 20.1 \\
Co-SLAM~\cite{wang2023coslam} & 1.00\tnote{h} & 30.2\tnote{h} & 8\tnote{e} & 1,258\tnote{e} & 125.0 \\
Point-SLAM~\cite{sandstrom2023point} & 0.52 & 35.2 & 2{,}865\tnote{e} & 6{,}563\tnote{e} & 0.7 \\
SplaTAM$^\dagger$~\cite{keetha2024splatam} & 0.36 & 34.1 & 254\tnote{e} & 14{,}024\tnote{e} & 10.9 \\
MonoGS$^\dagger$~\cite{matsuki2024gaussian} & 0.58 & 38.9 & 90\tnote{d} & ---\tnote{d} & 19.2 \\
GS-SLAM$^\dagger$~\cite{yan2024gsslam} & 0.50 & 34.3\tnote{h} & 198\tnote{c} & ---\tnote{d} & 10.1 \\
SGS-SLAM$^\dagger$~\cite{li2024sgsslam} & 0.41 & 34.7\tnote{h} & 254\tnote{e} & 40{,}330\tnote{e} & 9.6 \\
\bottomrule
\end{tabular*}
\begin{tablenotes}
\scriptsize
\item[a] Default width-256 network; 1.04\,MB per iMAP Table~2.
\item[b] Avg.\ ATE RMSE on Replica from Point-SLAM~\cite{sandstrom2023point} Table~1.
\item[c] Scene embedding for Replica Room0 from GS-SLAM~\cite{yan2024gsslam} Table~4.
\item[d] Survey estimate; not directly reported in the cited paper.
\item[e] Independently measured on Replica/room0 (NVIDIA A100-SXM4-80GB). Map size from saved checkpoint on disk; peak GPU memory via \texttt{nvidia-smi} at 1\,Hz, baseline subtracted.
\item[f] ATE for Room0 only from MonoGS~\cite{matsuki2024gaussian} Table~2; avg.\ across 8 Replica scenes is 2.58\,cm. Used here for consistency with single-scene profiling.
\item[g] Avg.\ PSNR on Replica from Point-SLAM~\cite{sandstrom2023point} Table~2 and MonoGS~\cite{matsuki2024gaussian} Table~5 (both report 24.42\,dB for NICE-SLAM).
\item[h] From GS-SLAM~\cite{yan2024gsslam}: ATE from Table~1, PSNR avg.\ 34.27\,dB from Table~6. SGS-SLAM PSNR from SGS-SLAM~\cite{li2024sgsslam} Table~1.
\end{tablenotes}
\end{threeparttable}
\end{table}

\subsection{Hierarchical and Semantic Representations}
\label{sec:hierarchical}

Topological maps grow with distinct places rather than volume, offering fundamentally different scaling. FabMap~\cite{cummins2008fabmap} pioneered appearance-based place recognition (topology only, no metric geometry). Dense semantic approaches such as SNI-SLAM~\cite{zhu2024snislam} jointly reconstruct geometry and semantics (ATE\,${\sim}$0.5\,cm on Replica). Hydra~\cite{hughes2022hydra} and its successors~\cite{zhang2025openfungraph} construct hierarchical scene graphs at 5\,Hz on a Jetson Xavier NX (8\,GB) with a 48\,MB graph abstraction, but this figure excludes the underlying Kimera~\cite{rosinol2020kimera} metric-semantic mesh (ATE\,${\sim}$5\,cm), making the total memory scene-dependent and underreported. Multi-robot extensions~\cite{tian2022kimeramulti} further multiply the footprint. The key memory question for scene graphs is \textit{which layers count}: the graph alone is compact, but the metric backend can rival a neural map.

Vision-language features introduce a new memory dimension. CLIP embeddings ($d{=}512$, 32-bit) cost ${\sim}2$\,GB per million points, comparable to the geometric map itself. Three strategies trade off memory against capability: HOV-SG~\cite{werby2024hovsg} stores features per segment (${\sim}75\%$ reduction), Clio~\cite{maggio2024clio} clusters by task relevance, and VLMaps~\cite{huang2023vlmaps} fuses features into a 2D grid. Downstream planning systems (LM-Nav~\cite{shah2023lmnav}, SayNav~\cite{rajvanshi2024saynav}, Open Scene Graphs~\cite{loo2024osg}, Embodied-RAG~\cite{xie2024embodiedrag}) inherit the map's memory profile and add planning-layer overhead that none quantify, making total system $\alpha$ unknown for semantic navigation.

\begin{table*}[t]
\caption{Unified efficiency analysis across spatial memory paradigms. Overhead factor $\alpha = M_{\text{peak}} / M_{\text{map}}$; $\alpha_{\text{CPU}}$ denotes CPU RSS, $\alpha_{\text{GPU}}$ denotes GPU allocation (see Section~\ref{sec:efficiency}). \textbf{Cross-benchmark caveat:} sparse systems are evaluated on EuRoC (real-world, stereo-inertial) and neural systems on Replica (synthetic, RGB-D); ATE and memory values are \textit{not} directly comparable across horizontal dividers. Within-benchmark comparisons are valid; cross-paradigm conclusions should be drawn from $\alpha$ and scaling behavior. Per-value provenance notes are in Tables~\ref{tab:feature_slam} and~\ref{tab:nerf_slam}.}
\label{tab:efficiency}
\centering
\scriptsize
\begin{threeparttable}
\begin{tabular*}{\textwidth}{@{\extracolsep{\fill}}llcccccc}
\toprule
\textbf{System} & \textbf{Paradigm} & \textbf{ATE (cm)} & \textbf{PSNR (dB)} & \textbf{Map (MB)} & \textbf{Peak (MB)} & \textbf{FPS} & \textbf{$\alpha$} \\
\midrule
\multicolumn{8}{l}{\textit{Sparse --- evaluated on EuRoC~\cite{euroc} (real-world, stereo-inertial)}} \\
\midrule
ORB-SLAM3~\cite{campos2021orb} & Sparse & 3.5 & --- & 55 & 220\tnote{$\dag$} & 30 & 4.0\tnote{$\dag$} \\
VINS-Mono~\cite{qin2018vins} & Sparse (VI) & 10.6\tnote{$\ddag$} & --- & 40 & --- & 20 & --- \\
Basalt~\cite{usenko2020basalt} & Sparse (VI) & 8.8\tnote{$\S$} & --- & 35\tnote{$\S$} & 120\tnote{$\S$} & 30\tnote{$\S$} & $\sim$3.4\tnote{$\S$} \\
\midrule
\multicolumn{8}{l}{\textit{Learned Flow --- evaluated on EuRoC / TUM RGB-D (real-world)}} \\
\midrule
DROID-SLAM~\cite{teed2021droidslam} & Learned Flow & 2.0\tnote{$\star$} & --- & ---\tnote{$\star\star$} & 11,000\tnote{$\star\star$} & 6 & --- \\
\midrule
\multicolumn{8}{l}{\textit{Neural --- evaluated on Replica~\cite{straub2019replica} (synthetic, RGB-D)}} \\
\midrule
iMAP~\cite{sucar2021imap} & NeRF & 3.12\tnote{$\star\star\star$} & 22.1 & 1 & --- & 3 & --- \\
NICE-SLAM~\cite{zhu2022nice} & NeRF & 1.06 & 24.4 & 47\tnote{$\diamond$} & 10,082\tnote{$\diamond$} & 2 & 215\tnote{$\diamond$} \\
Co-SLAM~\cite{wang2023coslam} & NeRF & 1.00 & 30.2 & 8\tnote{$\diamond$} & 1,258\tnote{$\diamond$} & 16 & 157\tnote{$\diamond$} \\
Point-SLAM~\cite{sandstrom2023point} & NeRF & 0.52 & 35.2 & 2{,}865\tnote{$\diamond$} & 6{,}563\tnote{$\diamond$} & 1 & 2.3\tnote{$\diamond$} \\
SplaTAM~\cite{keetha2024splatam} & 3DGS & 0.36 & 34.1 & 254\tnote{$\diamond$} & 14{,}024\tnote{$\diamond$} & --- & 55\tnote{$\diamond$} \\
MonoGS~\cite{matsuki2024gaussian} & 3DGS & 0.58 & 38.9 & 90 & --- & 3 & --- \\
GS-SLAM~\cite{yan2024gsslam} & 3DGS & 0.50 & 34.3 & 198 & --- & 8 & --- \\
SGS-SLAM~\cite{li2024sgsslam} & 3DGS (Sem.) & 0.41 & 34.7 & 254\tnote{$\diamond$} & 40{,}330\tnote{$\diamond$} & 2 & 159\tnote{$\diamond$} \\
\midrule
\multicolumn{8}{l}{\textit{Scale Reference --- evaluated on custom outdoor sequences (not directly comparable)}} \\
\midrule
GigaSLAM~\cite{deng2025gigaslam} & 3DGS (Hier.) & ---\tnote{$\|$} & $\sim$24\tnote{$\|$} & ---\tnote{$\|$} & --- & 3 & --- \\
\midrule
\multicolumn{8}{l}{\textit{Dense / Hierarchical --- hardware-dependent}} \\
\midrule
OctoMap~\cite{hornung2013octomap} & Dense (Octree) & --- & --- & 45 & --- & 10 & --- \\
Hydra~\cite{hughes2022hydra} & Scene Graph & --- & --- & 48\tnote{$\#$} & --- & 5 & --- \\
\bottomrule
\end{tabular*}
\begin{tablenotes}
\scriptsize
\item[$\ddag$] Average RMS ATE over MH01--MH05 (monocular-inertial); reproduced by~\cite{campos2021orb}, Table~II.
\item[$\S$] Survey estimates from profiling; ATE computed from Table~I of~\cite{usenko2020basalt} (stereo VIO, MH01--MH05).
\item[$\star$] Approximate average ATE on EuRoC (monocular) from \cite{teed2021droidslam} Table~2.
\item[$\star\star$] DROID-SLAM does not produce a persistent map file; $M_{\text{peak}}$ is GPU memory (requires $\geq$11\,GB per~\cite{teed2021droidslam}). $\alpha$ is not applicable.
\item[$\dag$] $\alpha_{\text{CPU}}$: CPU RSS measured on Intel i7-10700 (five runs, stereo-inertial).
\item[$\diamond$] $\alpha_{\text{GPU}}$: independently measured via \texttt{nvidia-smi} at 1\,Hz on Replica/room0; NVIDIA A100-SXM4-80GB. Baseline GPU memory subtracted. Map sizes from saved checkpoint files on disk. Checkpoint discrepancies (ours vs.\ prior literature): Co-SLAM 8 vs.\ 32\,MB, NICE-SLAM 47 vs.\ 235\,MB, Point-SLAM 2{,}865 vs.\ 80\,MB, SplaTAM 254 vs.\ 85\,MB, SGS-SLAM 254 vs.\ 92\,MB. See Section~\ref{sec:efficiency}.
\item[$\star\star\star$] Room0 only from MonoGS~\cite{matsuki2024gaussian} Table~2; avg.\ across 8 Replica scenes is 2.58\,cm.
\item[$\|$] GigaSLAM reports km-scale outdoor mapping; ATE is on custom sequences not directly comparable to Replica. PSNR is avg.\ on KITTI from Table~3. Map size not separately reported (GPU memory is 8--22\,GB including runtime overhead).
\item[$\#$] Graph abstraction layer only; excludes the underlying Kimera metric-semantic mesh, which adds additional memory.
\end{tablenotes}
\end{threeparttable}
\end{table*}

\section{Unified Efficiency Analysis}
\label{sec:efficiency}

A key contribution of this survey is a unified efficiency analysis comparing accuracy against memory cost \textit{across} paradigms. We structure this around two complementary metrics.

\subsection{The Overhead Factor \texorpdfstring{$\alpha$}{alpha}}

Published map sizes are misleading proxies for deployment cost. As illustrated by Co-SLAM~\cite{wang2023coslam} in Section~\ref{sec:neural}, a system saving an 8\,MB map requires 1.3\,GB of GPU memory at runtime. To capture this gap, we define the \textit{overhead factor}:
\begin{equation}
\alpha = \frac{M_{\text{peak}}}{M_{\text{map}}}
\label{eq:alpha}
\end{equation}
where $M_{\text{peak}}$ is peak runtime memory (CPU RSS or GPU allocation) and $M_{\text{map}}$ is the total persistent checkpoint size on disk (all files written by the system's save routine, including network weights, feature stores, and codebooks). This dimensionless ratio decomposes runtime cost into the map itself and the \textit{computational scaffolding}: optimizer state, gradient buffers, rendering caches, vocabulary data, and allocator overhead. Low $\alpha$ indicates that map size predicts deployment cost; high $\alpha$ indicates hidden runtime expenses. Note that $\alpha$ should always be interpreted alongside absolute $M_{\text{peak}}$ and $M_{\text{map}}$; Table~\ref{tab:efficiency} reports both quantities.

\textit{Important caveat:} We distinguish $\alpha_{\text{CPU}}$ (CPU RSS) from $\alpha_{\text{GPU}}$ (GPU device allocation). These are not directly comparable: CPU RSS includes libraries and OS overhead; GPU allocation captures optimizer state and CUDA caches but not the host process. Table~\ref{tab:efficiency} labels each explicitly. Within GPU methods alone, $\alpha_{\text{GPU}}$ spans from 2.3 (Point-SLAM, map-dominated) to 215 (NICE-SLAM, overhead-dominated).

Few systems report $M_{\text{peak}}$ directly. Our independent profiling reveals that $\alpha$ varies by two orders of magnitude: from ORB-SLAM3 ($\alpha_{\text{CPU}} = 4$, RSS on Intel i7-10700) and Point-SLAM ($\alpha_{\text{GPU}} = 2.3$, A100-SXM4-80GB) at one extreme, to NICE-SLAM ($\alpha_{\text{GPU}} = 215$) at the other, meaning a 47\,MB saved map consumes over 10\,GB of GPU memory. Critically, Point-SLAM's low $\alpha$ does not imply efficiency, since its 2.9\,GB checkpoint already dominates the 6.6\,GB peak, whereas Co-SLAM's low $M_{\text{map}}$ (8\,MB) masks 1.3\,GB of runtime cost. We advocate that future work routinely report $M_{\text{peak}}$ alongside $M_{\text{map}}$.

\textit{Architectural drivers of $\alpha$.} Three design choices largely determine $\alpha$: (1)~\textit{representation compactness vs.\ runtime scaffolding}: compact implicit methods (Co-SLAM, NICE-SLAM) store only network weights or hash tables on disk, but require optimizer state (Adam~\cite{kingma2015adam} stores two momentum buffers per parameter, ${\sim}3\times$ the model size), activation maps, and CUDA memory pool reservations at runtime, producing $\alpha \gg 1$; (2)~\textit{map-dominant architectures}: Point-SLAM stores per-point neural features directly in the checkpoint (2.9\,GB), so the map itself dominates runtime cost and $\alpha \approx 2.3$, but reducing $\alpha$ alone does not guarantee efficiency since Point-SLAM has the highest absolute $M_{\text{peak}}$ after SGS-SLAM; (3)~\textit{accumulation without pruning}: SplaTAM and SGS-SLAM continuously add Gaussians without merging or culling, amplifying $\alpha$ to 55--159 as rendering buffers scale with Gaussian count. In summary, low $\alpha$ can arise from genuinely efficient runtime (sparse SLAM) or from front-loading cost into the map (Point-SLAM); high $\alpha$ signals that runtime scaffolding dwarfs the persistent representation.

\textit{Measurement methodology.} We profiled five neural SLAM systems on Replica/room0 (NVIDIA A100-SXM4-80GB): Co-SLAM~\cite{wang2023coslam}, NICE-SLAM~\cite{zhu2022nice}, Point-SLAM~\cite{sandstrom2023point}, SplaTAM~\cite{keetha2024splatam}, and SGS-SLAM~\cite{li2024sgsslam}. Protocol: peak GPU memory sampled at 1\,Hz via \texttt{nvidia-smi}, baseline subtracted, map size from saved checkpoint. Checkpoint sizes revealed systematic discrepancies with literature in every case (ours vs.\ literature): Co-SLAM 8 vs.\ 32\,MB, NICE-SLAM 47 vs.\ 235\,MB, Point-SLAM 2{,}865 vs.\ 80\,MB, SplaTAM 254 vs.\ 85\,MB, SGS-SLAM 254 vs.\ 92\,MB. The 3DGS checkpoints are ${\sim}3\times$ larger because saved Gaussian parameters exceed typically reported counts. Values marked $\diamond$ in Table~\ref{tab:efficiency} are independently measured; Fig.~\ref{fig:memory_growth} shows memory time series.

\textit{Hardware sensitivity.} Much of the high $\alpha_{\text{GPU}}$ is training-time scaffolding (Adam~\cite{kingma2015adam} ${\sim}3\times$ overhead); inference-only deployment could substantially reduce $\alpha_{\text{GPU}}$, though no surveyed system profiles this. $\alpha$ is also hardware-dependent: Co-SLAM shows ${\sim}3.2$\,GB peak on RTX\,3090 (literature) vs.\ 1.3\,GB on A100 (measured), due to GPU-specific CUDA allocator behavior. We recommend that future $\alpha$ measurements report exact GPU model, baseline memory, and sampling method.

\subsection{Cross-Paradigm Comparison}

Table~\ref{tab:efficiency} presents the unified comparison. We omit composite efficiency ratios as they produce misleading rankings across benchmarks; the Pareto visualization (Fig.~\ref{fig:pareto}) preserves both dimensions. Within-benchmark comparisons are valid; cross-paradigm conclusions should be drawn from $\alpha$ and scaling behavior rather than absolute accuracy (see Table~\ref{tab:efficiency} caption).

\subsection{Pareto Analysis}

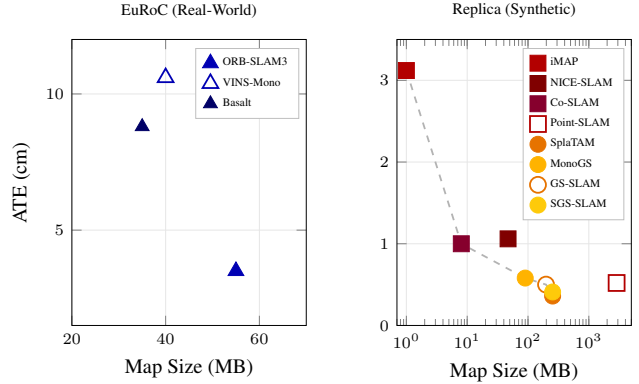
\begin{figure}[t]
\centering
\resizebox{\columnwidth}{!}{%
\begin{tikzpicture}
\begin{groupplot}[
    group style={group size=2 by 1, horizontal sep=1.4cm},
    xlabel={Map Size (MB)},
    ylabel={ATE (cm)},
    grid=major,
    grid style={gray!20},
    every axis label/.style={font=\small},
    tick label style={font=\scriptsize},
    width=5.2cm, height=6cm,
]
\nextgroupplot[
    title={\scriptsize EuRoC (Real-World)},
    xmin=20, xmax=70,
    ymin=1.5, ymax=12,
    legend style={at={(0.97,0.97)}, anchor=north east, font=\tiny, draw=gray!50, fill=white},
    legend cell align={left},
]
\addplot[only marks, mark=triangle*, mark size=4pt, color=blue!70!black] coordinates {(55, 3.5)};
\addlegendentry{ORB-SLAM3}
\addplot[only marks, mark=triangle, mark size=4pt, color=blue!70!black, thick] coordinates {(40, 10.6)};
\addlegendentry{VINS-Mono}
\addplot[only marks, mark=triangle*, mark size=3.5pt, color=blue!40!black] coordinates {(35, 8.8)};
\addlegendentry{Basalt}

\nextgroupplot[
    title={\scriptsize Replica (Synthetic)},
    xmode=log,
    xmin=0.7, xmax=5000,
    ymin=0, ymax=3.5,
    legend style={at={(0.97,0.97)}, anchor=north east, font=\tiny, draw=gray!50, fill=white, legend columns=1},
    legend cell align={left},
    ylabel={},
]
\addplot[only marks, mark=square*, mark size=3.5pt, color=red!70!black] coordinates {(1, 3.12)};
\addlegendentry{iMAP}
\addplot[only marks, mark=square*, mark size=3.5pt, color=red!50!black] coordinates {(47, 1.06)};
\addlegendentry{NICE-SLAM}
\addplot[only marks, mark=square*, mark size=3.5pt, color=purple!70!black] coordinates {(8, 1.00)};
\addlegendentry{Co-SLAM}
\addplot[only marks, mark=square, mark size=3.5pt, color=red!70!black, thick] coordinates {(2865, 0.52)};
\addlegendentry{Point-SLAM}
\addplot[only marks, mark=*, mark size=3.5pt, color=orange!90!black] coordinates {(254, 0.36)};
\addlegendentry{SplaTAM}
\addplot[only marks, mark=*, mark size=3.5pt, color=orange!60!yellow] coordinates {(90, 0.58)};
\addlegendentry{MonoGS}
\addplot[only marks, mark=o, mark size=3.5pt, color=orange!90!black, thick] coordinates {(198, 0.50)};
\addlegendentry{GS-SLAM}
\addplot[only marks, mark=*, mark size=3.5pt, color=yellow!60!orange] coordinates {(254, 0.41)};
\addlegendentry{SGS-SLAM}
\addplot[dashed, gray!60, thick, no marks] coordinates {
    (1, 3.12) (8, 1.00) (90, 0.58) (198, 0.50) (254, 0.36)
};
\end{groupplot}
\end{tikzpicture}%
}
\caption{Accuracy vs.\ map size ($M_{\text{map}}$) by benchmark. \textbf{Left:} Sparse (EuRoC, linear scale). \textbf{Right:} Neural (Replica, log scale); dashed = Pareto front. Lower-left is better. Point-SLAM's measured 2.9\,GB checkpoint (vs.\ 80\,MB survey estimate) moves it far right, off the Pareto front. Plotting $M_{\text{peak}}$ instead would further shift high-$\alpha$ systems rightward.}
\label{fig:pareto}
\end{figure}

Fig.~\ref{fig:pareto} visualizes the tradeoff landscape with explicit benchmark separation (EuRoC left, Replica right); these benchmarks are not directly comparable, hence both tables and figure separate them. Learned-flow systems (DROID-SLAM, DPVO) are excluded as they lack persistent maps. The dashed Pareto front on Replica traces five non-dominated points: iMAP$\to$Co-SLAM$\to$MonoGS$\to$GS-SLAM$\to$SplaTAM. The largest gain is iMAP$\to$Co-SLAM ($3\times$ ATE improvement at $8\times$ map cost); the three 3DGS systems then cluster in the 90--254\,MB range with diminishing returns (MonoGS map size is a survey estimate; GS-SLAM from their Table~4).

On EuRoC (Fig.~\ref{fig:pareto}, left), ORB-SLAM3 achieves the best ATE (3.5\,cm) at 55\,MB; Basalt trades accuracy for a smaller map (35\,MB). On Replica (right), 3DGS systems cluster favorably (ATE $\leq 0.58$\,cm, 90--254\,MB), with SplaTAM achieving the best ATE (0.36\,cm) at $\alpha_{\text{GPU}} = 55$. Point-SLAM's true checkpoint (2.9\,GB) places it far from the Pareto front, invisible under the 80\,MB survey estimate. The binding constraint differs by paradigm: runtime overhead for compact implicit methods ($\alpha_{\text{GPU}} = 157$, Co-SLAM) vs.\ map size for dense-point methods ($\alpha = 2.3$, Point-SLAM).

\section{Evaluation and Practical Guidance}
\label{sec:evaluation}

The overhead factor $\alpha$ captures an important dimension, but a single ratio is insufficient for characterizing spatial memory efficiency. We propose four complementary metrics that no surveyed system currently reports:

\begin{enumerate}
\item \textbf{Memory growth rate} ($dM/dt$): distinguishes bounded systems (RTAB-Map, hash-based NeRF) from unbounded ones (vanilla 3DGS); low final $M_{\text{map}}$ but high $dM/dt$ leads to out-of-memory failures on longer missions.
\item \textbf{Query latency}: varies by orders of magnitude ($O(1)$ hash lookups to full NeRF inference) yet is never benchmarked comparatively.
\item \textbf{Memory--completeness curve}: F1 score vs.\ cumulative map size, revealing diminishing returns in reconstruction quality.
\item \textbf{Throughput degradation}: FPS as map size approaches the memory ceiling; short benchmarks ($<$5\,min) never stress this.
\end{enumerate}

\begin{figure}[t]
\centering
\resizebox{\columnwidth}{!}{%
\begin{tikzpicture}
\begin{axis}[
    xlabel={Run Progress (\%)},
    ylabel={GPU Memory (MB)},
    grid=major, grid style={gray!20},
    width=9cm, height=5cm,
    every axis label/.style={font=\small},
    tick label style={font=\scriptsize},
    xmin=0, xmax=100,
    legend style={at={(0.5,-0.32)}, anchor=north, font=\scriptsize, legend columns=2, draw=gray!50, fill=white, column sep=6pt},
]
\addplot[blue!70!black, thick] coordinates {
    (5,1258) (10,1258) (20,1258) (30,1258) (40,1258) (50,1258)
    (60,1258) (70,1258) (80,1258) (90,1258) (100,1226)
};
\addlegendentry{Co-SLAM$^\diamond$}
\addplot[red!70!black, thick] coordinates {
    (5,3000) (10,5937) (20,2921) (30,4150) (40,7022) (50,5716)
    (60,6226) (70,6852) (80,6282) (90,6688) (100,4550)
};
\addlegendentry{NICE-SLAM$^\diamond$}
\addplot[orange, thick] coordinates {
    (5,1700) (10,3470) (20,6038) (30,7894) (40,9214) (50,10556)
    (60,11898) (70,12296) (80,12780) (90,12488) (100,12740)
};
\addlegendentry{SplaTAM$^\diamond$}
\addplot[purple!70!black, thick] coordinates {
    (5,4800) (10,5317) (20,4889) (30,5009) (40,5095) (50,5289)
    (60,5515) (70,5613) (80,5807) (90,5510) (100,5512)
};
\addlegendentry{Point-SLAM$^\diamond$}
\end{axis}
\end{tikzpicture}%
}
\caption{Runtime GPU memory on Replica/room0 (1\,Hz sampling, A100-SXM4-80GB, baseline subtracted). All four systems independently measured ($\diamond$). Co-SLAM stays flat at ${\sim}1.3$\,GB (bounded hash). NICE-SLAM oscillates between 3--7\,GB due to periodic global mapping passes (absolute peak 10.1\,GB, Table~\ref{tab:efficiency}). Point-SLAM stabilizes at 5--6\,GB after initialization. SplaTAM grows monotonically to ${\sim}12.7$\,GB as Gaussians accumulate (peak 14.0\,GB).}
\label{fig:memory_growth}
\end{figure}
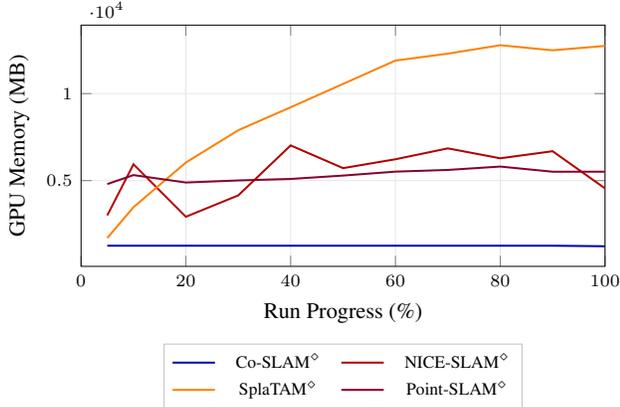

We demonstrate memory growth rate via independent profiling (Fig.~\ref{fig:memory_growth}): GPU memory sampled at 1\,Hz on Replica/room0, showing that Co-SLAM is bounded, SplaTAM grows linearly, and NICE-SLAM oscillates with periodic global passes. The remaining three metrics are not yet benchmarked in any surveyed system; we formalize them here as concrete targets for future evaluation infrastructure.

\textit{Proposed evaluation protocol.} We recommend that future work:
(1)~report $M_{\text{map}}$, $M_{\text{peak}}$, and $\alpha$ with hardware specs;
(2)~evaluate on sequences $\geq$30\,min to stress memory management;
(3)~separate training-time from inference-time $\alpha$;
(4)~profile on both discrete-GPU and embedded platforms.
Current benchmarks (EuRoC~\cite{euroc}, Replica~\cite{straub2019replica}, TUM RGB-D~\cite{tumrgbd}) cover minutes at room-scale; long-duration benchmarks recording $M(t)$ and FPS$(t)$ would be a high-impact contribution.

\subsection{Deployment Budgeting}
\label{sec:conclusions}

We distill our analysis into a systematic procedure. Given a target platform, Algorithm~\ref{alg:budgeting} computes the largest map the hardware can support; Table~\ref{tab:selection_guide} then narrows the paradigm choice. The key insight: work \textit{backward} from deployment memory constraints using $\alpha$ (Table~\ref{tab:efficiency}), not forward from map size benchmarks:
\begin{equation}
M_{\text{map}}^{\max} = \frac{M_{\text{budget}}}{\alpha}
\label{eq:budget}
\end{equation}

\begin{algorithm}[t]
\caption{$\alpha$-Aware Memory Budgeting}\label{alg:budgeting}
\KwIn{$M_{\text{budget}}$: available RAM/VRAM after OS and drivers}
\KwOut{Feasible map size and representation choice}
Look up $\alpha$ for candidate paradigm (Table~\ref{tab:efficiency})\;
$M_{\text{map}}^{\max} \leftarrow M_{\text{budget}} \;/\; \alpha$ \tcp*{Eq.~\ref{eq:budget}}
\eIf{$M_{\text{map}}^{\max} \geq M_{\text{target}}$}{
    Select representation from Table~\ref{tab:selection_guide}\;
}{
    Apply compression ($10$--$25\times$~\cite{lee2024compact,fan2024lightgaussian}), adopt submaps/streaming, or choose sparser representation\;
}
\end{algorithm}

\textit{Example:} Jetson Orin NX (16\,GB) with high-$\alpha$ neural SLAM ($\alpha_{\text{GPU}} \in [55, 215]$): $M_{\text{map}}^{\max} \approx 75$--$290$\,MB. With sparse SLAM ($\alpha_{\text{CPU}} \approx 4$): ${\sim}4$\,GB. These $\alpha$ values are room-scale; building-scale deployments may differ.

\begin{table}[t]
\caption{Representation selection guide by primary constraint.}
\label{tab:selection_guide}
\centering
\scriptsize
\resizebox{\columnwidth}{!}{%
\begin{threeparttable}
\begin{tabular}{lcccc}
\toprule
\textbf{Constraint} & \textbf{Recommended} & \textbf{$\alpha$ range} & \textbf{Type} & \textbf{Avoid} \\
\midrule
CPU-only ($<$8\,GB) & Sparse, Octree & 3--5\tnote{a} & CPU & Neural \\
Embedded GPU ($<$16\,GB) & Sparse, SG & 4--10\tnote{$*$} & CPU & Raw 3DGS \\
Dense geometry & TSDF, 3DGS & ${\sim}$55\tnote{b} & GPU & Sparse only \\
Photo rendering & 3DGS, NeRF & 2--215\tnote{c} & GPU & Occupancy \\
Multi-hour & Submaps, Stream & varies & --- & Monolithic \\
Semantic & SG, VLM & 4--10\tnote{$*$} & CPU & Geom.\ only \\
\bottomrule
\end{tabular}
\begin{tablenotes}
\scriptsize
\item[a] Basalt ($\alpha_{\text{CPU}} \approx 3.4$) to ORB-SLAM3 ($\alpha_{\text{CPU}} = 4.0$).
\item[b] SplaTAM ($\alpha_{\text{GPU}} = 55$, measured); hardware-dependent (Sec.~\ref{sec:efficiency}).
\item[c] Point-SLAM ($2.3$, measured) to NICE-SLAM ($215$, measured). SplaTAM ($55$), Co-SLAM ($157$), SGS-SLAM ($159$) in range.
\item[$*$] Graph layer only; total $\alpha$ with metric backend is higher.
\item[] \textit{Type}: CPU RSS or GPU allocation (Sec.~\ref{sec:efficiency}). SG = Scene Graph.
\end{tablenotes}
\end{threeparttable}%
}
\end{table}

\section{Memory Dynamics}
\label{sec:dynamics}

Even compact representations accumulate data without active management. We organize memory dynamics along three dimensions (Table~\ref{tab:dynamics_taxonomy}): \textit{update policy} (how new observations are incorporated), \textit{forgetting rule} (whether and how old data is discarded), and \textit{partitioning} (monolithic vs.\ hierarchical memory organization). A key distinction: forgetting reduces $M_{\text{map}}$ but not $\alpha$: Co-SLAM has no forgetting mechanism yet achieves bounded $M_{\text{map}}$ via its fixed-size hash table, while still exhibiting $\alpha = 157$ because runtime scaffolding is unchanged. Reducing $\alpha$ requires \textit{architectural} changes (inference-only deployment, gradient checkpointing, mixed-precision training) that shrink the scaffolding itself.

\begin{table}[t]
\caption{Memory dynamics taxonomy. Consol.\ = consolidation, Hier = hierarchical, Mono = monolithic. $\Delta M$ = reported memory reduction. $\alpha$ from Table~\ref{tab:efficiency} where available; note that forgetting reduces $M_{\text{map}}$ but not $\alpha$.}
\label{tab:dynamics_taxonomy}
\centering
\scriptsize
\begin{threeparttable}
\begin{tabular*}{\columnwidth}{@{\extracolsep{\fill}}lccccR{0.9cm}}
\toprule
\textbf{System} & \textbf{Update} & \textbf{Forget} & \textbf{Part.} & \textbf{$\alpha$} & \textbf{$\Delta M$} \\
\midrule
ORB-SLAM3~\cite{campos2021orb} & Incr+BA & Culling & Sub & 4 & ---\tnote{a} \\
RTAB-Map~\cite{labbe2019rtabmap} & Incr & WM$\to$LTM & Sub & --- & Bound.\tnote{b} \\
DSO~\cite{engel2018dso} & Window & Temporal & Mono & --- & --- \\
MS-SLAM~\cite{zhang2024msslam} & Window & Sparsif. & Mono & --- & $>$70\%\tnote{c} \\
Co-SLAM~\cite{wang2023coslam} & Incr & None & Mono & 157 & --- \\
GigaSLAM~\cite{deng2025gigaslam} & Incr & LOD & Hier & --- & LOD\tnote{d} \\
MemGS~\cite{bai2025memgs} & Incr & Merging & Mono & --- & ---\tnote{e} \\
BioSLAM~\cite{yin2023bioslam} & Gated & Consol. & Hier & --- & +24\%\tnote{f} \\
Hydra~\cite{hughes2022hydra} & Incr & None & Hier & --- & --- \\
\bottomrule
\end{tabular*}
\begin{tablenotes}
\scriptsize
\item[a] Culling impact not quantified in the original paper.
\item[b] Active memory bounded by WM size parameter regardless of session length.
\item[c] $>$70\% reduction in peak memory increase vs.\ ORB-SLAM3~\cite{zhang2024msslam}.
\item[d] Level-of-detail rendering reduces active Gaussians; total map persists on disk.
\item[e] Merging reduces Gaussian count but impact not separately quantified.
\item[f] +24\% weighted recall vs.\ generative replay; memory impact not reported.
\end{tablenotes}
\end{threeparttable}
\end{table}

Our profiling (Fig.~\ref{fig:memory_growth}) maps directly onto this taxonomy: Co-SLAM's flat trace reflects its pre-allocated hash (bounded $M(t)$), NICE-SLAM oscillates from periodic mapping spikes, SplaTAM grows monotonically without pruning (cf.\ MemGS~\cite{bai2025memgs}), and Point-SLAM stabilizes after initialization. All four neural systems lack explicit forgetting. Classical methods are more disciplined: ORB-SLAM3 culls keyframes, RTAB-Map transfers to disk, DSO~\cite{engel2018dso} applies temporal windowing, and BioSLAM~\cite{yin2023bioslam} uses gated consolidation (Table~\ref{tab:dynamics_taxonomy}).

Multi-robot systems add a distribution dimension: centralized~\cite{schmuck2019ccmslam,schmuck2021covins} vs.\ decentralized~\cite{lajoie2024swarmslam} architectures multiply $M_{\text{map}}$, and collaborative 3DGS~\cite{xu2025macego3d,yu2025hammer,thomas2025grandslam} is emerging with unknown merging cost~\cite{nguyen2025cogslamsurvey}.

\section{Open Challenges}
\label{sec:challenges}

Sparse systems exhibit \textit{explicit} failure modes (tracking loss triggers relocalization~\cite{qin2018vins}), whereas neural methods degrade \textit{silently} (hash saturation~\cite{wang2023coslam}) or \textit{catastrophically} (out-of-memory termination~\cite{bai2025memgs}).

\paragraph{City-scale neural SLAM.} GigaSLAM~\cite{deng2025gigaslam} demonstrates kilometer-scale 3DGS, but true city-scale operation (${\sim}10^7$\,m$^3$, $>$100\,h, $<$30\,W) remains open. Even at $15\times$ compression~\cite{fan2024lightgaussian}, city-scale maps reach multi-gigabyte range without LOD management. DiskChunGS~\cite{li2025diskchungs} addresses streaming via spatial chunking, but the embedded gap remains severe (${\sim}0.01$\,FPS on Jetson AGX Orin~\cite{gao2025semanticslamembedded}). Cloud-edge splitting is established for feature-based SLAM~\cite{benali2020edgeslam,schmuck2019ccmslam,patel2023covinsg} but unexplored for 3DGS.

\paragraph{Lifelong map maintenance.} Neural SLAM faces catastrophic forgetting~\cite{kirkpatrick2017overcoming}: new observations overwrite weights encoding earlier regions. Mitigations include elastic weight consolidation~\cite{kirkpatrick2017overcoming}, CL-Splats~\cite{ackermann2025clsplats}, VBGS~\cite{vandemaele2024vbgs}, GaussianUpdate~\cite{jeon2025gaussianupdate}, and WildGS-SLAM~\cite{zheng2025wildgsslam}, yet all target single-session neural map updates. BioSLAM~\cite{yin2023bioslam} addresses multi-session place recognition but not map-level continual learning for neural representations. Information-theoretic criteria for \textit{when to forget} (e.g., retaining only observations that reduce map entropy) are largely unexplored.

\paragraph{Uncertainty quantification.} Neural representations return confident outputs for unobserved regions, a critical failure mode for safety-critical planning. Stochastic approaches multiply inference cost, and storing per-element variance doubles map storage. Moment-based uncertainty~\cite{ewen2025magicmoments} propagates moments through rendering without additional training, though memory overhead in full SLAM is unverified.

\paragraph{Foundation model features.} CLIP features ($d{=}512$, 32-bit) cost ${\sim}2$\,GB per million points, adding a memory layer \textit{on top of} the geometric map. Compression strategies include per-segment storage (HOV-SG~\cite{werby2024hovsg}, 75\% reduction), task-driven clustering (Clio~\cite{maggio2024clio}), and product quantization (Dr.\ Splat~\cite{sun2025drsplat}, ${\sim}6\%$ ratio). Even compressed, language features add ${\sim}50\%$ per-Gaussian overhead. No surveyed paper reports $\alpha$ for semantic systems.

\paragraph{Hardware--algorithm co-design.} Unified memory architectures (NVIDIA Jetson Orin, Apple Silicon) may reduce $\alpha$ by eliminating duplicate CPU--GPU buffers. Custom ASICs (REACT3D~\cite{liu2025react3d}: $12\times$ throughput) and software optimization~\cite{wang2025vigsfusion} (14\,FPS on Orin NX) narrow the gap but neither achieves full-resolution real-time on stock hardware.

\paragraph{Toward scalable solutions.} No single technique permanently solves memory scaling, because data complexity grows with environment size and mission duration. However, three complementary strategies converge toward \textit{bounded-memory operation}: (1)~\textit{hierarchical streaming} (cf.\ GigaSLAM~\cite{deng2025gigaslam}, DiskChunGS~\cite{li2025diskchungs}), bounding runtime memory at the cost of disk I/O; (2)~\textit{information-theoretic forgetting}, achieving $O(\log t)$ growth in principle; and (3)~\textit{hardware--algorithm co-design}, reducing $\alpha$ structurally. A system combining all three would approach indefinite operation under fixed memory, where cost depends on the \textit{active working area} rather than total mission history. No existing system achieves this.

\section{Conclusion}

This survey traced spatial memory evolution across over three decades and 88 references around a central finding: \textit{map size alone is an unreliable proxy for deployment cost}. Our overhead factor $\alpha$ spans two orders of magnitude ($\alpha_{\text{GPU}} = 2.3$--$215$), revealing that compact implicit methods pay orders-of-magnitude runtime overhead, dense neural point methods embed cost in the map itself, and 3DGS methods inherit high $\alpha$ from unbounded accumulation. No single paradigm dominates (Fig.~\ref{fig:pareto}), and memory growth dynamics (Fig.~\ref{fig:memory_growth}) reveal failure modes invisible in static metrics.

\textit{Recommendations.} (1)~Adopt $\alpha$-aware reporting ($M_{\text{peak}}$ + hardware specs alongside $M_{\text{map}}$); (2)~develop long-duration benchmarks ($\geq$30\,min) that record $M(t)$ and stress-test growth to failure; (3)~establish inference-time profiling protocols separating training overhead from deployment cost.

\textbf{Limitations.} Our profiling covers five neural SLAM systems on a single Replica scene using one GPU (A100); $\alpha$ values may differ on other scenes, datasets, or hardware. Only training-time $\alpha$ is measured; inference-only $\alpha$ remains unprofiled. Benchmark heterogeneity constrains cross-paradigm comparisons to $\alpha$ and scaling behavior.

This survey is designed to be actionable: the taxonomy (Fig.~\ref{fig:taxonomy}) maps the landscape, Table~\ref{tab:selection_guide} matches paradigms to hardware, and Algorithm~\ref{alg:budgeting} computes maximum feasible map size from the memory budget and $\alpha$. As neural SLAM matures toward deployment, we expect $\alpha$-aware design to become standard practice.

\clearpage
{
    \small
    \bibliographystyle{ieeenat_fullname}
    \bibliography{main}
}

\end{document}